\documentclass[runningheads]{llncs}
\usepackage[T1]{fontenc}
\usepackage{graphicx,verbatim}
\usepackage{threeparttable}
\usepackage{multirow}
\usepackage{booktabs}
\usepackage{makecell}
\usepackage[table]{xcolor}
\usepackage{pifont}
\usepackage{array}
\usepackage{xcolor}
\usepackage{amsmath}
\usepackage{amssymb}

\begin{document}
\title{DeNuC: Decoupling Nuclei Detection and Classification in Histopathology}

\author{Zijiang Yang\inst{1} \and
Chen Kuang\inst{1} \and
Dongmei Fu\inst{1,2}}
\authorrunning{Yang et al.}
\institute{School of Automation and Electrical Engineering, University of Science and Technology Beijing, Beijing 100083, China \and
Beijing Engineering Research Center of Industrial Spectrum Imaging
\email{\{zijiangyang, ckuang\}@xs.ustb.edu.cn, fdm\_ustb@ustb.edu.cn}\\
}

\maketitle              
\begin{abstract}
Pathology Foundation Models (FMs) have shown strong performance across a wide range of pathology image representation and diagnostic tasks. However, FMs do not exhibit the expected performance advantage over traditional specialized models in Nuclei Detection and Classification (NDC). In this work, we reveal that jointly optimizing nuclei detection and classification leads to severe representation degradation in FMs. Moreover, we identify that the substantial intrinsic disparity in task difficulty between nuclei detection and nuclei classification renders joint NDC optimization unnecessarily computationally burdensome for the detection stage. To address these challenges, we propose \textbf{DeNuC}, a simple yet effective method designed to break through existing bottlenecks by \textbf{De}coupling \textbf{Nu}clei detection and \textbf{C}lassification. DeNuC employs a lightweight model for accurate nuclei localization, subsequently leveraging a pathology FM to encode input images and query nucleus-specific features based on the detected coordinates for classification. Extensive experiments on three widely used benchmarks demonstrate that DeNuC effectively unlocks the representational potential of FMs for NDC and significantly outperforms state-of-the-art methods. Notably, DeNuC improves F1 scores by 4.2\% and 3.6\% (or higher) on the BRCAM2C and PUMA datasets, respectively, while using only 16\% (or fewer) trainable parameters compared to other methods.
Code is available at https://github.com/ZijiangY1116/DeNuC.


\end{abstract}

\section{Introduction}

Nuclei Detection and Classification (NDC) is a cornerstone of quantitative analysis and diagnosis in computational pathology~\cite{diao2021human,zhang2025systematic}.
Recent efforts have shown promising results by exploring spatial context representation~\cite{mcspatnet}, nuclei graph~\cite{cgt}, or leveraging tissue-level context features~\cite{pathcontext}.
Despite these approaches generally validating that enhancing nuclear morphology and tissue context representation is key to improving performance, they often come at the cost of introducing highly complex module designs, which not only necessitate excessive design for specific datasets but also increase computational redundancy, thereby limiting the generalization of models across diverse clinical scenarios.

\begin{figure}[t]
    \centering
    \includegraphics[width=0.95\textwidth]{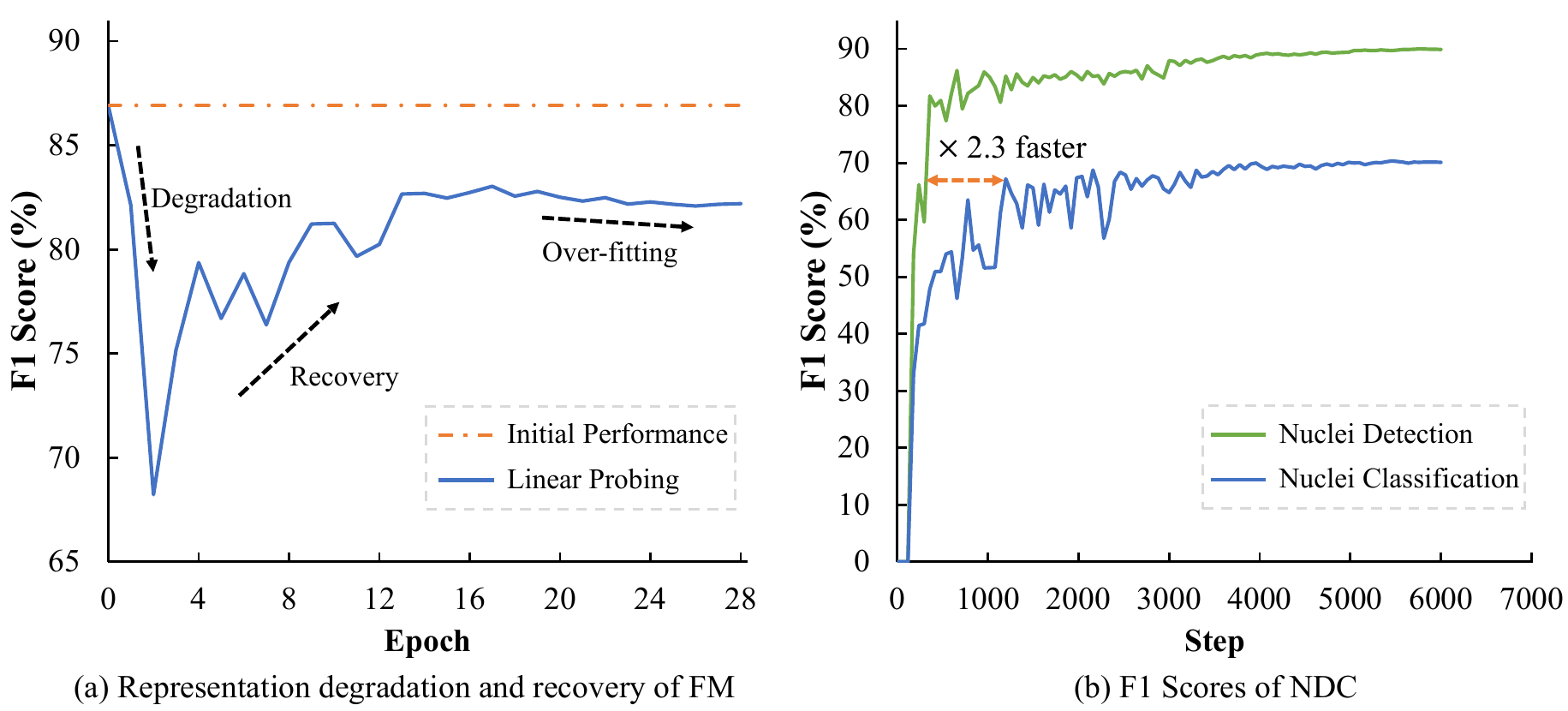}
    \label{fig:first_fig}
    \caption{
    Analysis of model performance during the NDC training.
    (a) We evaluate the representation capability of UNI2-H~\cite{uni} during joint optimization for NDC on the OCELOT~\cite{ryu2023ocelot} dataset via linear probing. UNI2-H undergoes a severe representation degradation in the early training phase. Although the performance subsequently recovers, it fails to regain its initial optimal level and rapidly deteriorates into over-fitting.
    (b) We evaluate the detection and classification F1 scores of a ConvNeXt-S~\cite{convnext} model pre-trained solely on ImageNet-1K throughout the training process on PUMA~\cite{puma}. Despite lacking domain-specific pathology pre-training, nuclei detection converges approximately 2.3 $\times$ faster than classification, highlighting the inherent difficulty disparity between the two tasks.
    }
\end{figure}

Recently, benefiting from a generic design and pre-training on large-scale unlabeled data, pathology Foundation Models (FMs)~\cite{uni,gigapath} exhibit strong general visual representation capabilities for pathology images, significantly improving performance on numerous downstream tasks~\cite{ctranspath}.
This has fostered the expectation that high-performance NDC can be achieved with simple architectures by directly leveraging the robust feature extraction of FMs.
However, compared to traditional specialized models, FMs have not demonstrated a significant performance advantage in nucleus-level tasks~\cite{cellvit++}.

In this work, we reveal that FMs suffer from severe representation degradation when jointly optimizing nuclei localization and classification.
Although pathology pre-trained models effectively encode semantic representations of images, they are not inherently designed for coordinate regression.
Forcing simultaneous joint optimization of classification and regression disrupts the pre-trained feature space, preventing the effective utilization of the original robust representations of FMs for NDC.
As illustrated in Fig.\ref{fig:first_fig} (a), although the FM initially possesses strong nuclei representation capabilities, the model parameters undergo updates during joint training to rapidly adapt to nuclei localization requirements.
While this improves detection performance, it causes a severe decline in the quality of pure nuclei representations.
In subsequent training stages, although classification representation capability recovers, it fails to return to its initial optimal level.

Furthermore, we identify a significant disparity in optimization difficulty between nuclei detection and nuclei classification, which exacerbates the inefficiency of joint optimization. Nuclei typically exhibit a relatively fixed size range and distinct contrast against the stromal background, resulting in lower optimization difficulty for detection.
As shown in Fig.\ref{fig:first_fig} (b), even a model without pathology pre-training can achieve high-performance nuclei detection within a very short training period, whereas nuclei detection converges approximately 2.3 $\times$ faster than classification.
This implies that formulating NDC as a multi-task joint optimization problem not only fails to achieve mutual complementarity between tasks but also significantly inflates the computational cost of detection.

\begin{figure}[t]
    \centering
    \includegraphics[width=0.95\textwidth]{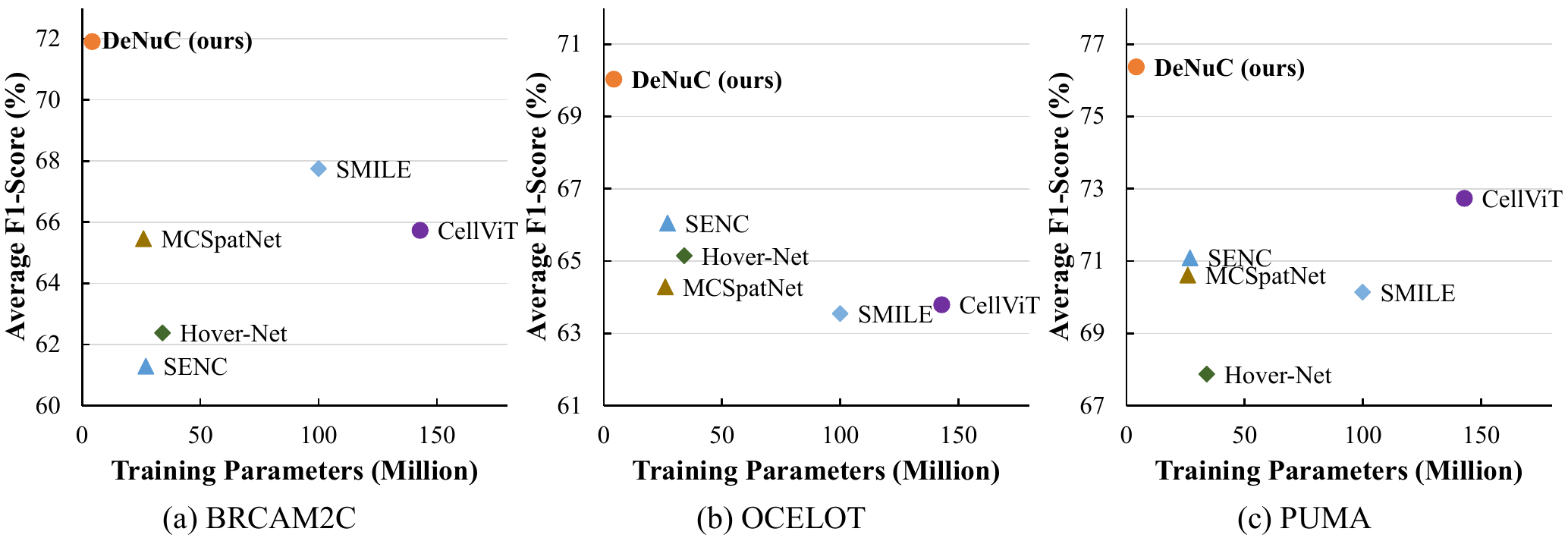}
    \caption{
    Comparison of DeNuC and SOTA methods.
    DeNuC achieves significantly superior performance across three benchmark datasets.
    \label{fig:compare_of_methods}
    }
\end{figure}

To address these issues, we propose a simple yet effective method named \textbf{DeNuC}, designed to break through existing bottlenecks by \textbf{De}coupling \textbf{Nu}clei detection and \textbf{C}lassification.
Specifically, DeNuC first employs a lightweight detection model to localize all nuclei in the input pathology image.
Subsequently, we leverage a pathology foundation model to encode the input image and utilize the detected coordinates to query nucleus-specific features from the feature maps for classification.
For nuclei detection, DeNuC not only compresses model parameters to an extreme minimum but also unlocks the capability for cross-dataset joint detection learning.
For nuclei classification, this decoupled design allows the foundation model to focus on fine-grained nuclei representation without being disturbed by localization task gradients, thereby fully unleashing its pre-trained representation potential.
As shown in Fig.\ref{fig:compare_of_methods}, extensive experiments on three widely used benchmarks indicate that DeNuC not only significantly outperforms existing State-Of-The-Art (SOTA) methods in performance but also requires only 16\% or less of the training parameters compared to other models.

\section{Methodology}

\subsection{Problem Formulation}

Given an input pathology image $X\in\mathbb{R}^{H\times W\times 3}$, the objective of NDC is to identify both the spatial localization and category of nuclei:
\begin{equation}
P = \mathcal{F}(\mathbf{X}) = \{ (\mathbf{p}_k, c_k) \}_{k=1}^{K},
\label{eq:ndc_definition}
\end{equation}
where $P$ is the set of predictions, $\mathcal{F}$ is the model, and $K$ denotes the total number of detected nuclei.
Each prediction consists of a centroid coordinate vector $\textbf{p}=(x_k, y_k) \in \mathbb{R}^2$ locating the nucleus in $X$, and a scalar $c_k \in \{1, \dots, C\}$ representing the predicted class category, with $C$ being the total number of defined nuclei types.
The set of nuclei coordinates $\{\textbf{p}_k\}_{k=1}^K$ and the set of nuclei types $\{c_k\}_{k=1}^K$ are denoted as $P_{det}$ and $P_{cls}$, respectively.

\subsection{DeNuC}

\noindent \textbf{Decoupling NDC.}
Existing methods typically formulate Equation (\ref{eq:ndc_definition}) as a multi-task problem that share the backbone:
\begin{equation}
     P = \{ \underbrace{\mathcal{H}_{det}(\mathcal{B}(\mathbf{X}))}_{\text{Detection}}, \quad \underbrace{\mathcal{H}_{cls}(\mathcal{B}(\mathbf{X}))}_{\text{Classification}} \},
\label{eq:ndc_joint_optim}
\end{equation}
where $\mathcal{B}$ denotes the backbone encoder, while $\mathcal{H}_{cls}$ and $\mathcal{H}_{det}$ represent the detection and classification modules, respectively.
In map-based methods~\cite{hovernet}, an additional decoder is required for nuclei segmentation~\cite{shui2024unleashing}. Moreover, $\mathcal{H}_{cls}$ and $\mathcal{H}_{det}$ are implemented as parameter-free post-processing algorithms.
Conversely, anchor-based methods~\cite{dpap2pnet,pathcontext,MUSE} implement them as lightweight Multi-Layer Perceptrons (MLPs) for regression and prediction.
Given that map-based methods are susceptible to morphological variations and necessitate heuristic tuning—thereby limiting performance, we focus our analysis primarily on the anchor-based methods to streamline the discussion.

In anchor-based frameworks, the backbone is required to extract feature maps that not only effectively represent nuclei appearance but can also be exploited by $\mathcal{H}_{cls}$ to regress the relative offsets between anchor points and potential ground-truth nuclei locations.
Although pathology pre-trained models excel at representing histopathological images, they are inherently unsuitable for coordinate regression.
Jointly optimizing NDC disrupts the pre-trained feature space, hindering the effective utilization of the robust representations derived from foundation models.

\begin{figure}
    \includegraphics[width=0.9\textwidth]{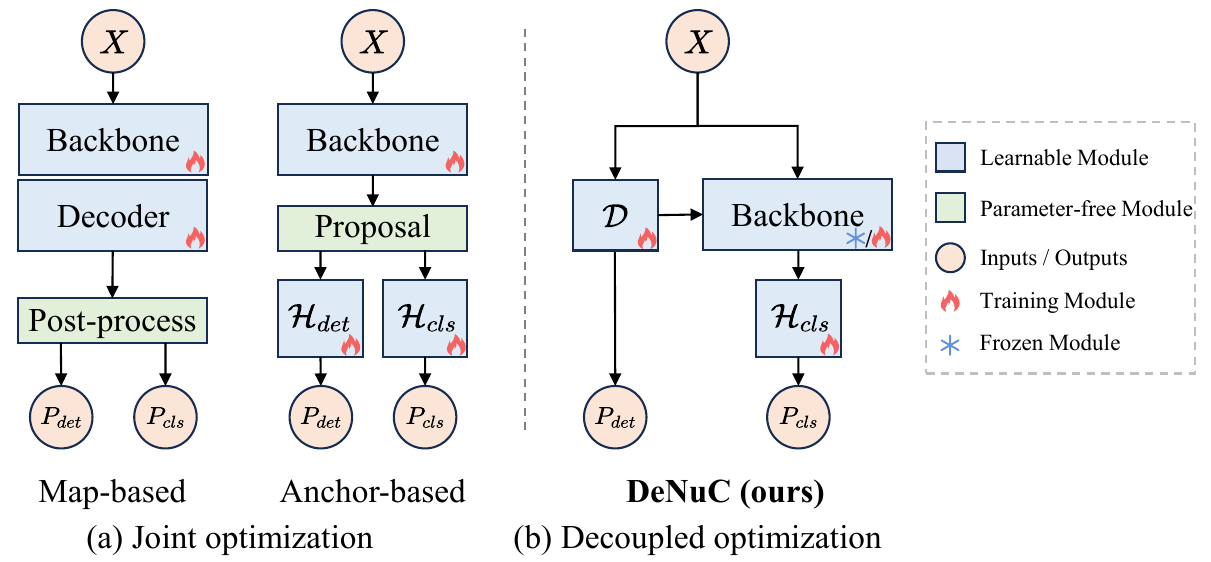}
    \caption{
    Illustration of (a) Joint optimization and (b) Decoupled optimization for NDC.
    Existing methods require the backbone to accommodate additional optimization objectives beyond nuclei representation, leading to representation degradation.
    In contrast, DeNuC employs an independent lightweight detection network $\mathcal{D}$ for nuclei localization, thereby allowing the backbone to focus exclusively on representation learning.
    }
    \label{fig:main_fig}
\end{figure}

As illustrated in Fig.\ref{fig:main_fig}, we address this bottleneck by decoupling detection and classification:
\begin{equation}
     P = \{ \underbrace{\mathcal{D}(\mathbf{X})}_{\text{Detection}}, \quad \underbrace{\mathcal{H}_{cls}(\mathcal{B}(\mathbf{X}, \mathcal{D}(\mathbf{X})))}_{\text{Classification}} \},
\label{eq:ndc_decouple}
\end{equation}
where $\mathcal{D}(X)$ represents an lightweight detection model. This formulation separates NDC into two distinct optimization objectives: (i) $\mathcal{D}(X)$ is dedicated solely to nuclei localization and (ii) $\mathcal{B}$ focus on extracting nucleus-specific features for classification, conditioned on the input image and the detected coordinates.
As demonstrated in our experiments, this decoupled framework not only fully exploits the powerful representation capabilities of pathology foundation models but also significantly alleviates the computational burden of detection.

\noindent \textbf{Nuclei Detection.}
To achieve efficient and robust nuclei localization, we adopt a single-stage point detection architecture inspired by P2PNet~\cite{p2pnet}.
Given an input pathology image $X$, the detection model $\mathcal{D}$ extracts feature maps and constructs a corresponding reference grid $\mathcal{G}$. For each spatial location $(i, j)$ on the grid, the network concurrently predicts a nuclei confidence score $s_{i, j}$ and a spatial coordinate offset $\boldsymbol{\delta}_{i,j}\in\mathbb{R}^2$ relative to the grid point. The final set of detected nuclei $P_{det}$ is then directly formulated as: 
\begin{equation}
P_{det} = \{ \textbf{g}_{i,j} + \boldsymbol{\delta}_{i,j} \mid s_{i,j} > \tau, \forall (i, j) \in \mathcal{G} \},
\label{eq:detection_module}
\end{equation}
where $\textbf{g}_{i,j}$ denotes the original coordinates of the grid point, and $\tau=0.5$ is the confidence threshold used to filter out background noise.
Crucially, Equation (\ref{eq:detection_module}) only necessitates binary classification between nuclei and background, thereby enabling the use of an extremely lightweight model for detection and facilitating joint training across multiple datasets.

\noindent \textbf{Nuclei Classification.}
Based on the detected nuclei locations $P_{det}$, the classification module further predicts the specific category for each nucleus. First, a pre-trained foundation model $\mathcal{B}$ extracts high-dimensional semantic feature maps $F\in\mathbb{R}^{C\times H^\prime \times W^\prime}$ of $X$.
Subsequently, to precisely capture the local context of each nucleus, we employ a bilinear interpolation sampling operation $\mathcal{S}$ to directly query the corresponding feature vectors from $F$ using the coordinate set $P_{det}$. The final classification prediction set $P_{cls}$ is obtained via $\mathcal{H}_{cls}$:
\begin{equation}
P_{cls} = \{ \mathcal{H}_{cls}(\textbf{f}_k) \mid \mathbf{f}_k = \mathcal{S}(F, \mathbf{p}_k), \forall \mathbf{p}_k \in P_{det} \},
\label{eq:classification_module}
\end{equation}
where $\textbf{p}_k$ denotes the k-th detected nuclei, and $\textbf{f}_k$ represents the sampled feature vector specific to that nucleus.
This coordinate-guided feature querying mechanism not only eliminates redundant computations but also ensures a high alignment between classification features and the spatial locations of nuclei.

\begin{table*}[t]
  \caption{
  Comparison of nucleus detection and classification in F1-score \% ($\uparrow$). $F^{Tum.}$, $F^{Lym.}$, $F^{Oth.}$, and $F^{Avg.}$ denote the F1 score of tumor nucleus, lymphocytes, other nucleus, and average, respectively. The best results are highlighted in \textbf{bold}, and the second-best results are in \underline{underlined}. DeNuC significantly outperforms other methods.}
  \centering
  \small
 \renewcommand\tabcolsep{2pt}
 \renewcommand{\arraystretch}{1.05}
 \resizebox{\textwidth}{!}{
 \begin{tabular}{cc|cccc|ccc|cccc}
    \toprule

    \multicolumn{1}{c}{\multirow{2}{*}{Method}} &
    \multicolumn{1}{c|}{Training} &
    \multicolumn{4}{c|}{BRCAM2C} & 
    \multicolumn{3}{c|}{OCELOT} & 
    \multicolumn{4}{c}{PUMA}\\
    & Params. & $F^{Lym.}$ & $F^{Tum.}$ & $F^{Oth.}$ & $F^{Avg.}$ & $F^{Tum.}$ & $F^{Oth.}$ & $F^{Avg.}$ & $F^{Lym.}$ & $F^{Tum.}$ & $F^{Oth.}$ & $F^{Avg.}$ \\

    \midrule

    CGT~\cite{cgt} & 37M & 56.42 & 75.98 & 50.44 & 60.95 & 68.77 & 61.30 & 65.03 & 76.55 & 79.66 & 54.20 & 70.14 \\
    SENC~\cite{senc} & 27M & 57.94 & 76.50 & 49.42 & 61.29 & 70.02 & 62.08 & 66.05 & 77.38 & 81.51 & 54.38 & 71.09 \\
    CellViT~\cite{cellvit} & 143M & 67.20 & 78.20 & 51.81 & 65.73 & 67.36 & 60.22 & 63.79 & 79.07 & 81.16 & \underline{57.96} & \underline{72.73} \\
    Hover-Net~\cite{hovernet} & 34M & 62.31 & 75.25 & 49.58 & 62.38 & 69.27 & 61.03& 65.15 & 75.72 & 78.36 & 49.53 & 67.87\\
    DPA-P2PNet~\cite{dpap2pnet} & 32M & 59.65 & 77.26 & \underline{55.26} & 64.06 & 70.07 & 59.92 & 64.99 & 76.80 & 81.87 & 54.04 & 70.90 \\
    MCSpatNet~\cite{mcspatnet} & 26M & 63.15 & 78.56 & 54.66 & 65.46 & 68.60 & 59.99 & 64.29 & 78.25 & \underline{82.05} & 51.54 & 70.61\\
    PointNu-Net~\cite{pointnu} & 160M & \underline{71.51} & 76.02 & 51.95 & 66.50 & 66.72 & 56.96 & 61.84 & 76.31 & 79.57 & 52.68 & 69.52 \\
    SMILE~\cite{smile} & 100M & \textbf{72.59} & \underline{79.61} & 51.06 & \underline{67.75} & 66.99 & 60.10 & 63.55 & \underline{80.35} & 77.54 & 52.52 & 70.14 \\
    MUSE~\cite{MUSE} & 123M & - & - & - & - & \underline{73.48} & \underline{64.52} & \underline{69.00} & - & - & - & - \\

    \midrule
    
    \textbf{DeNuC (ours)} & 4.3M & 69.73 & \textbf{85.10} & \textbf{61.08} & \textbf{71.97} & \textbf{73.83} & \textbf{66.04} & \textbf{69.94} & \textbf{81.00} & \textbf{85.25} & \textbf{62.85} & \textbf{76.37}\\
    
    \bottomrule
  \end{tabular}
  }
  \label{tab:finetune}
\end{table*}

\noindent \textbf{Learning.}
We optimize DeNuC via a two-stage training paradigm.
First, $\mathcal{D}$ is trained using a combination of L2 regression loss and binary cross-entropy loss.
Unless otherwise specified, $\mathcal{D}$ is jointly trained across multiple datasets to maximize the utility of all available annotations.
Then, $\mathcal{D}$ serves as an auxiliary network to facilitate classifier training. The classifier is optimized with standard cross-entropy loss. To maximize training efficiency, we freeze the backbone $\mathcal{B}$ and only optimize the classification head $\mathcal{H}_{cls}$, unless stated otherwise.
As demonstrated in our experiments, optimizing solely $\mathcal{D}$ and $\mathcal{H}_{cls}$ is sufficient to achieve SOTA performance.

\section{Experiments}

\subsection{Experiment Settings}

\noindent \textbf{Datasets and evaluation metrics.}
To comprehensively evaluate the performance of DeNuC, we conduct extensive experiments on three widely used public datasets, including BRCAM2C~\cite{mcspatnet}, OCELOT~\cite{ryu2023ocelot}, and PUMA~\cite{puma}.
Following the common practice of NDC~\cite{MUSE,pathcontext}, we employ the distance-based F1-score for evaluation.
Specifically, we perform a one-to-one matching between the predicted and ground-truth (GT) nuclei within the same category.
A prediction is identified as a True Positive (TP) if it successfully matches a GT nucleus within a predefined distance threshold.
Unmatched predictions, or those exceeding the distance threshold, are counted as False Positives (FP), while unmatched GT nuclei are designated as False Negatives (FN).
Furthermore, we report the average F1-score across all classes to assess the overall performance on the dataset.

\begin{table*}[t]
  \caption{
  Ablation study of $\mathcal{D}$ in F1-score \% ($\uparrow$). \textbf{(Left)} Comparison of different backbones. \textbf{(Right)} Ablation of cross-dataset training strategy. $N_{nu}$ denotes the number of nuclei in the training set. $F^{Det.}$ denotes the F1 score of detection. "SN" denotes ShuffleNetV2. "Separated" and "Joint" indicate that models are trained individually on each dataset and jointly across all datasets, respectively.}
  \centering
  \small
  \renewcommand{\arraystretch}{1.05}
  
  \begin{minipage}[c]{0.53\textwidth}
    \centering
    \renewcommand\tabcolsep{2pt}
    \resizebox{\linewidth}{!}{
    \begin{tabular}{cc|cccccc}
        \toprule
        \multicolumn{1}{c}{\multirow{2}{*}{Backbone}} & 
        \multicolumn{1}{c|}{\multirow{2}{*}{Params.}} & 
        \multicolumn{2}{c}{BRCAM2C} &
        \multicolumn{2}{c}{OCELOT} &
        \multicolumn{2}{c}{PUMA} \\
        & & $F^{Det.}$ & $F^{Avg.}$ & $F^{Det.}$ & $F^{Avg.}$ & $F^{Det.}$ & $F^{Avg.}$ \\
        \midrule
        SN (0.5$\times$) & 0.3M & 86.61 & 71.43 & 79.89 & 68.85 & 92.23 & 75.45 \\
        SN (1.0$\times$) & 1.0M & 86.98 & 71.58 & 80.97 & 69.74 & 93.13 & 75.98 \\
        SN (1.5$\times$) & 2.7M & 87.00 & 71.90 & 81.07 & 69.76 & 93.28 & 76.19 \\
        SN (2.0$\times$) & 4.3M & 87.52 & 71.97 & 81.33 & 69.94 & 93.57 & 76.36\\
        ResNet-50 & 26M & 87.29 & 71.90 & 81.48 & 70.03 & 93.22 & 76.07 \\
        \bottomrule
    \end{tabular}
    }
  \end{minipage}
  \hfill
  \begin{minipage}[c]{0.43\textwidth}
    \centering
    \renewcommand\tabcolsep{2pt}
    \resizebox{\linewidth}{!}{
    \begin{tabular}{cc|cccc}
        \toprule
        \multicolumn{1}{c}{\multirow{2}{*}{Dataset}} &
        \multicolumn{1}{c|}{\multirow{2}{*}{$N_{nu}$}} &
        \multicolumn{2}{c}{Separated} & 
        \multicolumn{2}{c}{Joint}\\
        & & $F^{Det.}$ & $F^{Avg.}$ & $F^{Det.}$ & $F^{Avg.}$ \\
        \midrule
        BRCAM2C & 18.6k & 85.15 & 70.26 & 87.52 & 71.97\\
        OCELOT & 65.8k & 81.22 & 69.90 & 81.33 & 69.94\\
        PUMA & 56.9k & 93.42 & 76.25 & 93.57 & 76.36\\
        \bottomrule
    \end{tabular}
    }
  \end{minipage}
  
  \label{tab:ablation_study_detection}
\end{table*}

\noindent \textbf{Baselines.}
We compare DeNuC with SOTA methods, including SENC~\cite{senc}, CellViT~\cite{cellvit}, Hover-Net~\cite{hovernet}, CGT~\cite{cgt}, DPA-P2PNet~\cite{dpap2pnet}, MCSpatNet~\cite{mcspatnet}, SMILE~\cite{smile}, and MUSE~\cite{MUSE}.
For fair comparison, we cite the results of MUSE evaluated at the same input size as the other models, instead of using LFoV samples.

\noindent \textbf{Implementation.}
$\mathcal{D}$ employs a pre-trained ShuffleNetV2~\cite{shufflenet} as the backbone and a Path Aggregation Network (PAN)~\cite{PAN} as the FPN. The detector is trained for 100 epochs, utilizing a batch size of 32 and an initial learning rate of 0.001.
UNI2-H~\cite{uni} is employed as the feature extractor.
$\mathcal{H}_{cls}$ is implemented as a single fully connected layer.
The classifier is trained for 100 epochs with a batch size of 256 and an initial learning rate of 0.01.

\subsection{Main Results}

Table \ref{tab:finetune} reports the comparative results for nuclei detection and classification. DeNuC demonstrates significant performance advantages over both existing map-based and anchor-based approaches.
Specifically, DeNuC outperforms current SOTA methods on BRCAM2C, OCELOT, and PUMA by substantial margins of 4.2\%, 0.9\%, and 3.6\% in $F^{Avg.}$, respectively. 
Notably, DeNuC achieves these superior results with only 4.3M trainable parameters, which constitutes 16\% or less of the parameters required by other methods. The results clearly demonstrate that DeNuC not only substantially improves the performance of nuclear analysis but also achieves remarkably high parameter efficiency.

\subsection{Ablation Studies}

\noindent \textbf{Detection model.}
Table \ref{tab:ablation_study_detection} reports the ablation results for the detection module $\mathcal{D}$.
Table \ref{tab:ablation_study_detection} (Left) shows that ShuffleNetV2 (0.5$\times$) with only 0.3M parameters achieves a detection F1-score highly comparable to that of the 26M ResNet-50. This suggests that model capacity has a negligible impact on nuclei detection, further verifying the low optimization difficulty of this task.
In addition, Table \ref{tab:ablation_study_detection} (Rights) shows that joint multi-dataset training effectively boosts performance on datasets with fewer samples.
Specifically, on BRCAM2C, joint training increases the detection F1-score by 2.4\%, which subsequently improves classification performance by 1.7\%.

\noindent \textbf{Optimization strategy of the classification network.}
Table \ref{tab:ablation_study_classification} reports the ablation study on the optimization strategy for the classification network. The results indicate that providing $P_{det}$ to allow the classifier to focus solely on nuclei representation significantly improves F1-scores. Notably, compared to end-to-end training, simply freezing the backbone $\mathcal{B}$ and optimizing only the linear classification head yields substantial gains of 3.9\%, 7.2\%, and 1.4\% on BRCAM2C, OCELOT, and PUMA, respectively.
Furthermore, optimizing $\mathcal{B}$ via LoRA or full fine-tuning leads to additional performance improvements.
These results strongly suggest that simultaneous end-to-end detection and classification induces representation degradation, whereas DeNuC effectively capitalizes on the robust visual representation capabilities of FMs for pathology images.

\begin{table*}[t]
  \caption{
  Ablation study on the optimization strategy of the classification network in F1-score \% ($\uparrow$). Providing nuclei coordinates highly improves performance.}
  \centering
  \small
 \renewcommand\tabcolsep{2pt}
 \renewcommand{\arraystretch}{1.05}
 \resizebox{\textwidth}{!}{
 \begin{tabular}{ccc|cccc|ccc|cccc}
    \toprule

    \multicolumn{1}{c}{\multirow{2}{*}{Method}} &
    \multicolumn{1}{c}{\multirow{2}{*}{$P_{det}$}} &
    \multicolumn{1}{c|}{Training} &
    \multicolumn{4}{c|}{BRCAM2C} & 
    \multicolumn{3}{c|}{OCELOT} & 
    \multicolumn{4}{c}{PUMA}\\
    & & Params. & $F^{Lym.}$ & $F^{Tum.}$ & $F^{Oth.}$ & $F^{Avg.}$ & $F^{Tum.}$ & $F^{Oth.}$ & $F^{Avg.}$ & $F^{Lym.}$ & $F^{Tum.}$ & $F^{Oth.}$ & $F^{Avg.}$ \\

    \midrule

    Linear & \ding{51} & 4.6K & 69.73 & 85.10 & 61.08 & 71.97 & 73.83 & 66.04 & 69.94 & 81.00 & 85.25 & 62.85 & 76.37 \\
    Lora &  \ding{51} & 2.4M & 71.79 & 85.33 & 61.40 & 72.84 & 74.19 & 66.93 & 70.56 & 81.23 & 85.70 & 63.38 & 76.77 \\
    Fully &  \ding{51} & 681M & 71.03 & 85.50 & 61.11 & 72.55 & 73.87 & 65.52 & 69.69 & 81.45 & 85.37 & 63.13 & 76.65 \\
    End-to-end & \ding{55} & 686M & 68.21 & 79.42 & 56.67 & 68.10 & 67.37 & 58.15 & 62.76 & 79.50 & 83.19 & 62.20 & 74.97 \\
    
    \bottomrule
  \end{tabular}
  }
  \label{tab:ablation_study_classification}
\end{table*}

\section{Conclusion}

In this work, we reveal that FMs suffer from severe representation degradation when jointly optimizing nuclei localization and classification.
Furthermore, we identify a significant disparity in task difficulty between nuclei detection and classification, implying that formulating NDC as a multi-task joint optimization problem not only fails to achieve mutual complementarity but also unnecessarily inflates the computational cost of detection.
To address these challenges, we propose \textbf{DeNuC}, a simple yet effective method designed to break through existing bottlenecks by \textbf{De}coupling \textbf{Nu}clei detection and \textbf{C}lassification.
Extensive experiments demonstrate that DeNuC significantly outperforms existing methods while requiring only 16\% or fewer training parameters compared to existing methods.
These results validate DeNuC as a new paradigm that combines high performance with high training efficiency, offering a novel perspective for developing high-precision and resource-efficient algorithms in NDC.

\bibliographystyle{splncs04}
\bibliography{mybibliography}

\end{document}